# Assessing Large Language Models in Mechanical Engineering Education: A Study on Mechanics-Focused Conceptual Understanding

Jie Tian[1*], Jixin Hou[1*], Zihao Wu[2*], Peng Shu[2], Ning Liu[3], Zhengliang Liu[2], Yujie Xiang[3], Beikang Gu[1], Nicholas Filla[1], Yiwei Li[2], Xianyan Chen[4], Keke Tang[3#], Tianming Liu[2#], and Xianqiao Wang[1#]

[1]School of Environmental, Civil, Agricultural and Mechanical Engineering, College of Engineering, University of Georgia, Athens, GA, 30602, USA

[2]School of Computing, University of Georgia, Athens, GA, 30602, USA

[3]School of Aerospace Engineering and Applied Mechanics, Tongji University, Shanghai, China

[4]Department of Epidemiology and Biostatistics, College of Public Health, University of Georgia, Athens, GA 30602, USA

[*]Co-first authors

[#]Corresponding authors: kktang@tongji.edu.cn, tliu@uga.edu, xqwang@uga.edu

**Abstract**

This study is a pioneering endeavor to investigate the capabilities of Large Language Models (LLMs) in addressing conceptual questions within the domain of mechanical engineering with a focus on mechanics. Our examination involves a manually crafted exam encompassing 126 multiple-choice questions, spanning various aspects of mechanics courses, including Fluid Mechanics, Mechanical Vibration, Engineering Statics and Dynamics, Mechanics of Materials, Theory of Elasticity, and Continuum Mechanics. Three LLMs, including ChatGPT (GPT-3.5), ChatGPT (GPT-4), and Claude (Claude-2.1), were subjected to evaluation against engineering faculties and students with/without mechanical engineering background. The findings reveal GPT-4's superior performance over the other two LLMs and human cohorts in answering questions across various mechanics topics, except for Continuum Mechanics. This signals the potential future improvements for GPT models in handling symbolic calculations and tensor analyses. The performances of LLMs were all significantly improved with explanations prompted prior to direct responses, underscoring the crucial role of prompt engineering. Interestingly, GPT-3.5 demonstrates improved performance with prompts covering a broader domain, while GPT-4 excels with prompts focusing on specific subjects. Finally, GPT-4 exhibits notable advancements in mitigating input bias, as evidenced by guessing preferences for humans. This study unveils the substantial potential of LLMs as highly knowledgeable assistants in both mechanical pedagogy and scientific research.

1. Introduction

Mechanical engineering stands as the forefront of technological advancement, serving as the cornerstone for innovation across diverse industries [1-11]. Yet, the study of fundamental mechanics, such as Mechanics of Materials, Theory of Elasticity, Engineering Statics and Dynamics, present various challenges for students due to their inherent complexities. Mechanics involves abundant abstract concepts that are not readily observable, making it challenging to conceptualize the terminologies like forces, torques, and moments. Additionally, the field heavily relies on mathematical expressions, hindering a holistic understanding of the underlying physical principles and significantly dampening enthusiasm for both student learning and educator teaching of mechanics. Meanwhile, traditional pedagogical methods often struggle to effectively bridge the gap between theoretical principles and practical applications, leaving students grappling with those intricate descriptions of material deformations and balance laws. This disparity underscores the need for innovative teaching methodologies that can enrich the learning experience and enhance comprehension in mechanical engineering education.

In the rapidly evolving landscape of Natural Language Processing (NLP), the emergence of Large Language Models (LLMs) signifies a transformative paradigm shift in artificial intelligence capabilities [12]. NLP, dedicated to facilitating computer-human language interaction, encompasses various applications ranging from language translation, content summarization to sentiment analysis [13]. Conventional NLPs are constrained in offering tailored solutions for specific applications through supervised training [14]. In this paradigm, models undergo fine-tuning on labeled training data, and often require further pre-training on domain-specific datasets to achieve optimal performance. In contrast, LLMs, characterized by their expansive size and exceptional few-shot learning capabilities, transcend these limitations by introducing the in-context learning approach [15-17]. The overarching impact of this method is a novel and streamlined NLP workflow, eliminating the need for supervised fine-tuning along with associated intricacies such as hyper-parameter tuning and model architecture modifications. LLMs has demonstrated significant potential in advancing scientific research across diverse domains, including biomedical research [18-20], material science [21, 22], chemistry [23-25], and environmental science [26]. For instance, Lee, et al. [27] show that BioBERT model outperforms state-of-art counterparts in biomedical text mining tasks, such as terminology recognition, relation extraction, and question answering. Another study by Jablonka, et al. [28] highlight the promising applications of LLMs in predicting properties of molecules and materials, designing interfaces, and extracting knowledge from unstructured data in chemistry and material science. In the realm of mechanics, while the use of LLMs is rather surprising, some noteworthy endeavors have been explored. Brodnik, et al. [29] point out that LLMs exhibit emergent capabilities in the area of applied mechanics, such as parallel information extraction, representations normalization to enrich downstream training dataset, programming assistance in computational and experimental mechanics, and hypothesis generation. A more in-depth exploration is provided by Buehler's group [30, 31], where a human-machine interactive MechGPT model, trained on a set of 1,103 Wikipedia articles related to mechanics, show satisfying potential in solving forward and inverse problem in various areas, including bio-inspired hierarchical honeycomb design, carbon nanotube mechanics, and protein unfolding.

A prominent example of LLMs is ChatGPT, developed by OpenAI. The first commercial version, GPT-3.5, was released in November 2022. Trained on an extensive internet dataset, this 175-billion-parameters LLMs showcases unprecedented proficiency in text generation, language translation, creative content creation, and furnishing informative response to a diverse array of queries [32]. Its successor, GPT-4, launched in March 2023, boasts an enhanced architecture and a more extensive dataset. It surpasses its predecessor in both size and capability, yielding outputs that are more cohesive, contextually relevant, and nuanced. Furthermore, GPT-4 integrates a robust multimodal capability, which allows it seamlessly incorporate information from various modalities such as text and images. GPT-4 has demonstrated an exceptional performance on diverse academic benchmarks. For example, it successfully passes simulated exams designed for humans, including the Uniform Bar Examination (MBE+MEE+MPT), Graduate Record Examination (GRE), and SAT Math [33]. However, it is worth noting that these exams boast substantial annual participation, and their resourceful testing materials are readily available on the internet. This suggests that the exam questions are probably included in the training data of GPT models [34]. Therefore, the superior performance in these examinations may be attributed to the extensive exposure to similar questions during training. An intriguing question arises: how does GPT perform in more specialized domain with smaller available training data, such as mechanical engineering? Addressing this query will provide insights into the potential of LLMs in revolutionizing academic and educational fields. While analogous efforts have been made in areas like radiation oncology [34] and synthetic biology [35], to the best of our knowledge, the capabilities of LLMs in solving mechanical problems remain unexplored.

In this manuscript, we embark on a pioneering journey, delving into the potential of LLMs to revolutionize the teaching and learning of fundamental mechanics within the mechanical engineering program. Following the GPT-4 technical report, we manually designed 126 multiple-choice questions pertaining to six fundamental mechanics disciplines: Fluid Mechanics, Mechanical Vibration, Engineering Statics and Dynamics, Mechanics of Materials, Theory of Elasticity, and Continuum Mechanics. Three state-of-art transformer-based LLMs were selected for evaluation: GPT-3.5, GPT-4, Claude-2.1. These results were compared with those obtained from human test-takers, including both experts and non-experts. As we navigate this intersection of LLMs and mechanics education, our objective is to harness the unique capabilities of LLMs to cultivate an enriched and interactive learning environment, ultimately reshaping the pedagogical landscape of the engineering mechanical program.

## 2. Related Work in LLMs

*2.1. Large language models*

Claude-2.1 [36] is the latest model developed by Anthropic, which delivers advancements in key capabilities for enterprises including an industry-leading 200K token context window, significant reductions in rates of model hallucination. Claude-2.1 distinguishes itself from other large language models by prioritizing ethical AI design and user safety. At the heart of Claude-2.1's design is a commitment to ethical AI principles. The model is engineered to minimize biases and ensure safe interactions. This makes Claude-2.1 particularly suited for applications where trust and reliability are crucial. Tool use is a new beta feature added in Claude-2.1 to integrate with users' existing processes, products, and APIs. This expanded interoperability aims to make Claude more useful across users' day-to-day operations. Claude-2.1 has also made significant gains in honesty, with a 2X decrease in false statements compared to the previous Claude-2.0 model. It promises to deploy AI across users' operations with greater trust and reliability.

Developed as an iteration of the revolutionary GPT-3 by OpenAI, GPT-3.5-Turbo-1106 model stands out for its specialized enhancements in processing speed and efficiency. It is the latest GPT-3.5 Turbo model with improved instruction following, JSON mode, reproducible outputs, parallel function calling, and more, which accepts up to 16,385 tokens for context window and returns a maximum of 4,096 output tokens. The "Turbo" in its name underscores its design philosophy – to provide rapid, high-performance language processing capabilities without compromising the depth and versatility for which the GPT series is renowned. This optimization is pivotal for applications requiring real-time interaction, such as conversational AI, where latency can significantly impact user experience. Despite its focus on speed, GPT-3.5-Turbo-1106 retains the comprehensive language understanding and generation capabilities characteristic of the GPT-3 model.

GPT-4 [33] is the next generation model in the groundbreaking Generative Pre-trained Transformer series by OpenAI, representing a monumental leap in the field of artificial intelligence and natural language processing. As a successor to the widely acclaimed GPT-3, GPT-4 continues to push the boundaries of what AI can achieve in understanding and generating human language. GPT-4 is a large multimodal model (accepting image and text inputs, emitting text outputs) that, while less capable than humans in many real-world scenarios, exhibits human-level performance on various professional and academic benchmarks. For example, it passes a simulated bar exam with a score around the top 10% of test takers; in contrast, GPT-3.5's score was around the bottom 10%. It is believed that GPT-4 consists of about 1.76 trillion parameters allowing this model even more versatile in various applications, from content creation, educational tools, and creative writing to more complex tasks like programming assistance and data analysis. With its advanced capabilities, GPT-4 also incorporates improved safety features and ethical considerations. OpenAI has aimed to mitigate risks associated with AI-generated content, such as misinformation and biases, making GPT-4 a more responsible and trustworthy AI model. GPT-4 marks a significant milestone in AI development. It not only demonstrates the rapid advancements in machine learning and AI capabilities but also highlights the increasing importance of ethical considerations in AI development.

*2.2. Language model examination*

Large language models exhibit exceptional natural language comprehension capabilities and are trained on vast datasets, endowing them with extensive knowledge. These attributes render large language models as ideal candidates for academic and professional benchmarking.

OpenAI recently published the inaugural study in the literature evaluating large language models on academic and professional exams intended for educated humans. This study reveals that GPT-4 excels across a broad spectrum of subjects, from the Uniform Bar Exam to the GRE. Additionally, a Microsoft study demonstrates GPT-4's ability to pass the USMLE, a professional exam for medical residents, with a significant margin.

Researchers have applied plenty of benchmarks to test the performance of their large models. For example, General Language Understanding Evaluation (GLUE) [37] is a collection of nine different tasks designed to test a model's ability to understand English language. These tasks include sentiment analysis, textual entailment, and question answering. It assesses the model's understanding of grammar, logic, and the relationships between sentences. SuperGLUE [38] is an extension of the GLUE benchmark but with more challenging tasks including reading comprehension, word sense disambiguation, and more complex question-answering challenges. Stanford Question Answering Dataset (SQuAD) [38] is a dataset consisting of questions posed by crowdworkers on a set of Wikipedia articles. The answers to these questions are segments of text from the corresponding reading passage. It tests a model's reading comprehension ability and how well it can extract answers from a given text. The LAMBADA [39] dataset evaluates the capabilities of language models in text understanding and prediction, particularly focusing on predicting the final word in a passage of text. These examinations provide a comprehensive overview of a model's linguistic abilities, ranging from basic understanding to advanced reasoning and comprehension.

*2.3. Prompt engineering*

The collection and labeling of data for training or fine-tuning NLP models can be resource-intensive and expensive, particularly in specialized fields like medicine. Recent research indicates that large-scale pre-trained language models (PLMs) can be adapted to downstream tasks using prompts, thereby potentially circumventing the need for fine-tuning.

A prompt is a set of instructions that tailor or refine the LLM's response, extending beyond simple task description or output format specification. Indeed, they can be engineered to facilitate novel interactions. For instance, ChatGPT can be prompted to simulate a cybersecurity breach using fictitious terminal commands. Furthermore, prompts can be used to generate additional prompts through a self-adaptation process.

The emergence of prompt engineering marks the beginning of a new era in natural language processing. The potential for diverse and complex applications of well-crafted prompts is undeniable. However, the challenge lies in determining the ideal prompt in the era of large language models. Currently, prompts can be manually created or automatically generated. While automatically generated prompts may excel in certain tasks, they often lack human readability and explainability. Thus, in domains where interpretability is paramount, such as in education and research, manual prompt generation may be preferable. In this study, we develop a series of prompts and chain-of-thought prompts, informed by our expertise and cooperated with GPTs develop kit provided ChatGPT website.

## 3. Methodology and Strategy

For this investigation, an evaluation consisting of 126 multiple-choice questions pertinent to mechanical engineering program with a focus on mechanics was formulated by a ground of seasoned experts in the field. This assessment covers six distinct domains: Fluid Mechanics, with 16 questions; Mechanics of Vibration, 20 questions; Engineering Statics and Dynamics, 24 questions; Mechanics of Materials, 25 questions; Theory of Elasticity, 19 questions; and Continuum Mechanics, 22 questions. The specific questions featured in this examination are comprehensively detailed in Supplemental Material.

*3.1. Question design*

As illustrated in Supplemental Material, the majority of the questions in the assessment predominantly focus on definitions and factual knowledge, alongside a considerable number of calculation-based questions. To more thoroughly comprehend the design of the examination (depicted in Figure 1), an analysis of the distribution of correct answers was conducted. Figure 1a showed that options B and C were more frequently selected as the correct answers, aligning with a common human intuition that places these options "in the middle." Figure 1b further delineates the distribution of correct answers across each domain, indicating that for each domain, three options follow a near-normal distribution pattern, while one option is consistently less favored.

*3.2. Strategy for comparison between LLM scores and human scores*

<u>Human Test Data Collection</u>

To formulate a thorough benchmark, examination results were collated from a diverse array of individuals. After an extensive review process, which involved the exclusion of evidently problematic results (such as those with incorrect lengths or options not listed in the question), data were collected from four distinct human test groups. This data set encompasses results from 17 undergraduate students majoring in non-mechanical engineering disciplines, 14 undergraduate students majoring in mechanical engineering, 9 graduate students in mechanics (inclusive of both Master's and Ph.D. students), and 6 engineering faculty members. For a detailed visualization of this data, please see Figure 2. The examination format for each human participant was consistent, allowing no time restrictions, enforcing a closed-book policy, and permitting the use of a basic calculator.

<u>Large Language Models Evaluated and Hyperparameter Setting</u>

This study is dedicated to the evaluation of two extensively utilized large language models (LLMs), specifically OpenAI and Claude, with no preceding inputs. For assessing OpenAI, the GPT-4 chat box serves as a benchmark, and their latest models, GPT-3.5-turbo-1106 and GPT-4-1106-preview, are tested with a temperature setting of 1 and a maximum token value of 4,096. Claude's most recent iteration, Claude-2.1, is evaluated under a hyperparameter configuration of temperature $T = 1$, $top_k = 40$, and $top_p = 0.9$.

To ensure the dependability of responses from LLMs, each question is subjected to multiple trials with requests for varied responses. As detailed in Table 1, specific prompts are utilized for different trials. These prompts are categorized according to their function: System prompts, which establish the role of the LLM, and Question prompts, which determine whether the LLM should first clarify the question before answering. We have developed three types of system prompts: 1. Simple prompts, general in nature, guiding LLMs to respond to scientific inquiries; 2. Mechanics-specific prompts, which direct the LLMs to tackle Mechanics questions and clearly refer to all six domains; 3. Domain-specific prompts, explicitly highlighting the pertinent domain of the question.

To create a benchmark for LLM performance, a single trial was conducted using Chatbox (GPT-4), incorporating domain-specific prompts and necessitating explanations. Both GPT-4 and GPT-3.5 were subjected to six trials, each encompassing three rounds. In these rounds, the question options were shuffled, and the OpenAI models were directed to produce five responses in each instance. For Claude-2.1, a similar trial was conducted, mirroring the approach used with Chatbox (GPT-4) and utilizing the identical prompt. This trial included three rounds, with each round yielding three responses.

Score Calculation and Evaluation

The test scores of large language models (LLMs) and their distribution patterns were compared both among themselves and with those of human cohorts. In this comparative analysis, the mean scores, consistency of scores, and confidence in responses were key evaluation metrics. To quantify the overall consistency in scoring success, we computed the standard deviation and the average correlation between trials. This average correlation, defined as the mean of the upper values in the Pearson correlation matrix across trials, assesses the uniformity of correct scores between trials. A value of 1 signifies identical distributions, 0 indicates a completely random distribution, and -1 represents a perfect anti-correlation. Additionally, to measure the confidence level in the responses provided by the LLMs and human groups, we tallied the number of correct answers for each question across all trials.

*3.3. Strategy for ChatGPT accuracy improvement*

Recent studies have found that transformer-based language models can greatly enhance response accuracy by generating answers in a stepwise manner. This is made possible by the LLM's ability to predict the following word based on prior context. To achieve this, it is necessary to prompt a sufficiently large LLM to generate responses step by step. In our research, we have improved the prompt by explicitly indicating the domain of questions or requesting the LLM to explain its reasoning process before providing an answer. Table 1 showcases various types of prompts that were utilized. For the OpenAI model, the domain-specific aspect is included in the system prompt, while the explanation prompt is in the user prompt. It is important to note that, for domain-specific prompts, the derived step pertains to all domains before being further refined into a specific domain.

Owing to the inherent design of transformer-based LLMs, which predict subsequent words based on preceding context, it has been demonstrated that the accuracy of responses can be enhanced if a sufficiently expansive LLM is prompted to construct answers in a step-wise fashion [40]. This approach was applied to ChatGPT (GPT-4) to investigate whether its performance could be augmented by prompting it to first elucidate and then respond. To refine the outcomes, we compared the results across trials where the question prompt either directed the LLM to provide an answer directly or to first explain, then answer, as depicted in each row of Table 1.

*3.4. Reasoning capability of ChatGPT*

In transformer-based models, the preceding output serves as the input for generating the subsequent token, forming the foundation of prompt engineering and chain-of-thought reasoning [40]. Specifically, ChatGPT and GPT-4 have demonstrated significant advancements in reasoning across various scenarios, such as solving mathematical and meme understanding. These LLMs excel in inductive, abductive, and analogical reasoning, which rely on pattern recognition and retrieval from pretrained knowledge, followed by correlating and adapting to a given set of observations to deduce conclusions. While these conclusions are insightful, they are not guaranteed to be accurate. Such reasoning is effective for general domain questions, where tasks involve directly extracting or summarizing answers from the provided context by utilizing their extensive pre-training knowledge. In contrast, applications in mechanical engineering may require LLMs to engage in deductive reasoning. This reasoning type involves drawing conclusions based on the truth of the premises, thus enabling the deduction of solid conclusions through a rigorous logical chain. To further assess ChatGPT's reasoning abilities, particularly in terms of human-like thinking and reasoning, we analyzed the similarity in accuracy distribution between human cohorts and LLMs.

## 4. Results

*4.1. Comparison between LLM scores and human scores*

Figure 2 delineates the raw scores for both human participants and LLMs. The LLM results are domain-specific and emphasize the importance of providing explanations first. Figure 3 highlights the mean test scores, revealing that contemporary models such as GPT-4 and Claude-2.1 demonstrate superior performance compared to engineering faculty. It is noteworthy that engineering faculty, rather than human experts, were chosen for comparison, as they often have expertise in one or several specific areas. Considering the closed-book nature of the exam, their scores might not fully represent the expertise typically expected. Generally, AI models appear to surpass non-expert human performance. The raw scores of undergraduate students majoring in non-mechanical engineering fields, with a mean accuracy near 25%, suggest a performance level akin to random guessing, corroborated by the distribution patterns shown in Figure 4.

In Figures 4b and 4c, the accuracy of engineering faculty and various LLMs across different domains is presented. These results indicate that answers accompanied by explanations generally lead to higher accuracy. Notably, GPT-4 shows a significant improvement in performance over GPT-3.5. Continuum Mechanics emerges as the domain where human engineering faculty maintain a notable edge over LLMs. This phenomenon can be attributed to the massive symbolic representations and mathematic calculations in the question design, especially those related to the tensor analysis.

The comparative analysis depicted in Figure 4b and Figure 5 suggests that the accuracy distribution models of engineering faculty and various large language models, such as Claude, GPT-3.5 (answer only), and GPT-4 (answer only), are similar. However, when the models are prompted to provide both an answer and an explanation, the accuracy distribution aligns more closely with human results. This implies that increased reasoning enhances the credibility and confidence in AI-generated responses. As shown in Figure 5c, we measure the correlation between trials from both Engineering Faculty and LLMs. Since human results are collected from different persons, The correlation is significantly lower, and LLMs all have greater confidence. Especially GPT-4 with explanation provided first.

*4.2. Comparison on the prompt engineering*

Based on the data presented in Figure 4d, restricting the domain before posing questions can significantly enhance the accuracy of the AI model. Furthermore, we can examine the differences between various domains and evaluate the significance of providing explanations, as shown in Figure 6. The technique of prompt engineering reveals improved answer performance after an explanation. For GPT-3.5, we can conclude that the prompt covering all domains has a higher likelihood of yielding optimal results. Regarding GPT-4, a more specific domain typically leads to a higher accuracy, but the improvement is not as substantial as with GPT-3.5.

*4.3. Guessing preference – option distribution of incorrect answers*

According to Figure 1, a tendency exists among humans to favor certain answers while setting up exam questions. Similarly, when confronted with unfamiliar questions, human respondents often lean towards specific choices, as evidenced in Figure 7a, where incorrect responses predominantly cluster around options B and C, even though the correct answer lies nearby. To delve deeper, we analyzed the option preference of LLMs, which exhibited no input bias due to the randomization of question options. Figure 7b elucidates that in the coding scheme, "E" signifies that GPT models identify none of the options as correct, "F" indicates that all options are correct, and "G" denotes a failure in generation. The distribution patterns suggest that GPT-3.5 shows a predilection for options A and C, whereas GPT-4 does not display a discernible preference for any specific option.

*4.4. Reasoning capability of ChatGPT*

When comparing the accuracy distribution of responses between engineering faculty and the GPT model, a clear pattern emerges: incorporating explanations significantly increases the randomness of incorrect answers and bolsters the confidence in correct ones. As illustrated in Figure 7b, responses generated without accompanying explanations tend to include some non-existent answers. While this might impair overall performance, prioritizing the provision of explanations before answers is essential for augmenting the model's capacity for deductive reasoning through the questions. We further assessed the reasoning capabilities of LLMs by examining the congruence in accuracy distribution between human cohorts and LLMs, as depicted in Figures 5c and 5d. For the correlation computation, we initially subtracted the mean value to mitigate bias stemming from the correct answer target. Incorporating explanations into LLM responses seemingly augments their resemblance to human-like attributes.

## 5. Discussions

The objective of this research is to evaluate the capabilities of large language models (LLMs) within the specific context of mechanical engineering with a focus on mechanics. A comprehensive 126-question examination, devised by a group of experts in the field (refer to the Supplemental Material), was employed to gauge the proficiency of LLMs. This assessment primarily targeted OpenAI models, namely GPT-3.5 and GPT-4, and included a comparative analysis with Claude-2.1. The findings reveal that all tested LLMs outperformed non-expert human participants. Notably, GPT-4 achieved higher scores than engineering faculty members. However, in more advanced and knowledge-intensive areas, such as Continuum Mechanics, human experts maintained a superior edge over LLMs.

The study's focus on prompt engineering significantly augmented the evaluation of LLM performance. Three types of prompt systems were identified, each aiding in the precise categorization of question domains. While GPT-4 exhibited some enhancements, its proficiency appears somewhat constrained. In contrast, GPT-3.5 demonstrated a more pronounced efficacy, particularly within the mechanics domain. However, this effectiveness varied when the focus was narrowed to specific areas like Fluid Mechanics. The OpenAI models, particularly when prompted to provide explanations before responses, displayed a high level of reasonability. The shift from moderate to either low or high accuracy in the results, compared to faculty and Claude's accuracy distribution, suggests an increase in confidence and precision. ChatGPT (GPT-4) showcased exceptional performance, with prospects for future advancements. The decision to exclude a custom-tailored GPT assistant from testing was based on concerns over potential overfitting. Overall, the study underscores the significant potential for collaboration between mechanics experts and ChatGPT (GPT-4) as a highly proficient assistant.

A limitation in assessing LLMs through exams, as demonstrated in this study, is that such exams predominantly reflect textbook knowledge, restricting their applicability mainly to educational contexts. Consequently, the relative performance of LLMs compared to engineering faculty in mechanics exams might not accurately represent the parity between LLMs and individual expertise in mechanics. Moreover, the high performance of GPT-4 in this certificate-like exam, focusing on a highly specialized subject, raises concerns about the depth of knowledge evaluated. This leads to a conundrum: if taken at face value, it could imply that GPT-4 possesses sufficient competence to function as a medical physicist, which appears highly unlikely. Thus, the mechanics community might need to reconsider their certificate procedures, as the need for humans to extensively learn such potentially superficial knowledge might decrease with the ongoing evolution of LLMs. In this context, LLMs could serve as a benchmark for superficiality in knowledge assessment. Emphasizing knowledge areas that LLMs have not mastered might become increasingly important.

*5.1. Applying large language models in mechanical engineering*

The incorporation of large language models emerges as a groundbreaking asset in the domains of mechanical engineering education and scientific research. For example, LLM-based chatbots, such as ChatGPT, can serve as dynamic and interactive tutors, providing real-time assistance to students in learning complex mechanical concepts with clarity and engagement. Moreover, LLMs like PaLM exhibits promising proficiency in tasks such as information extraction, synthesis, and distillation across diverse fields. They excel in activities ranging from code generations and translations between programming languages to guiding experiments and explaining methodologies [29]. However, understanding the scope and limitations of the language model is essential both for learners and educators in mechanical engineering [41, 42]. Primarily, the reliability and accuracy of these models should be meticulously assessed, i.e., verifying their information sources and relevance to mechanics domain. On the other hand, a thorough evaluation of the ethical implications is paramount, especially concerning academic integrity and responsible use of technology. Last, we should be mindful of the potential bias in the training corpus and seek diverse perspectives to ensure a comprehensive and well-rounded understanding of mechanics.

The finetuning of LLMs offers a powerful strategy for bolstering the capability of scientific research. In the research context, finetuning enables the customization of pre-trained language models to align with the specific nuances and terminologies inherent in the mechanical engineering field. For example, Luu and Buehler [43] developed the BioinspireLLM through finetuning with a corpus comprising over a thousand peer-reviewed articles in structural biology and bio-inspired materials. Their model exhibited remarkable capabilities in formulating sound hypotheses regarding biological materials design and predicting novel materials, thereby reshaping the traditional materials design process. A similar endeavor is the MechGPT [30], which was finetuned on a set of 1,103 Wikipedia articles related to mechanics. MechGPT demonstrates exceptional performance in solving both forward and inverse problems.

*5.2. Multi-modal models in mechanical engineering*

Multi-modal models, representing the forefront of language model development, hold significant potential in the field of mechanics. By integrating capabilities such as image recognition, these models could play a pivotal role in addressing complex mechanics problems. For instance, in the realm of structural engineering, multi-modal models can analyze visual data from structural inspections alongside textual reports and sensor readings to assess the integrity and safety of bridges or buildings. This comprehensive analysis enables more accurate predictions of potential structural failures or maintenance needs, leading to safer and more efficient management of infrastructure. Furthermore, in aerodynamics, multi-modal models could analyze airflow patterns over a vehicle's surface captured through visual methods, correlate this data with computational fluid dynamics simulations, and adjust design parameters accordingly. Additionally, certain reinforcement learning systems, such as AutoGPT, show promise in aiding structural design. In essence, multi-modal models in mechanics are paving the way for more integrated, intelligent, and predictive approaches to mechanical challenges. As these models continue to evolve, they are poised to become indispensable tools in the mechanical and structural engineering fields.

## 6. Conclusion

Our examination of large language models (LLMs) and human participants in mechanical engineering assessments reveals compelling insights. LLMs, notably GPT-4, demonstrate superior performance over human cohorts in closed-book exams related to engineering mechanics, and exhibit remarkable capabilities in deductive reasoning. The incorporation of explanations in prompts design consistently enhances both answering accuracy and deductive reasoning for LLMs. Prompt engineering proves critical for enhancing model accuracy, with GPT-3.5 excelling in a broader domain and GPT-4 preferring more specific domains. Additionally, the exploration of guessing preferences highlights human biases, while LLMs like GPT-4 demonstrate advancements in mitigating input bias. Overall, these findings suggest a promising future for LLMs in mechanical education and scientific research settings, paving the way for their increased integration and impact in these domains.

**Data Availability Statement**

The original contributions presented in the study are included in the article/supplemental material. Further inquiries can be directed to the corresponding authors.

**Conflict of Interest**

The authors declare that the research was conducted in the absence of any commercial or financial relationships that could be construed as a potential conflict of interest.

**Acknowledgements**

We would like to thank the many individuals (currently undergraduate students, graduate students, and engineering faculty in both Tongji University, China and University of Georgia, USA) that volunteered to take the mechanical engineering assessment.

# Figures & Captions

Table 1: **Prompt of different trials.** This table comprises three rows and two columns. Each cell includes two distinct prompts: the first paragraph functions as the system prompt, and the second paragraph signifies the beginning of the user's prompt. For example, in the OpenAI API, the initial paragraph is the "system prompt," and the subsequent paragraph forms the "user prompt." In other large language models (LLMs), these prompts may be amalgamated. Variations across rows primarily stem from the first paragraph, which may range from a basic to a mechanics-specific or a domain-specific prompt. Distinctively, the second paragraph separates the columns: the first column contains prompts necessitating an explanation, while the second column includes prompts that require LLMs to provide answers without an explanation.

| Prompts | With explanation | Answer Only |
| --- | --- | --- |
| Simple Prompt | You are a professor in mechanics, designed to help answer mechanics questions, providing expert knowledge in physics and engineering. It can explain concepts, solve problems, and offer insights into mechanical systems. You should exactly follow the instruction to provide the answer.<br><br>Can you answer the following single choice question? Provide the explanation, show the detailed steps of solving the problem in explanation section. Then answer the question by giving the choice, In the answer include the letter of the choice and the option content. For example, if the answer is A, then you should answer "A: option A content". You answer should follow the give format of question. | You are a professor in mechanics, designed to help answer mechanics questions, providing expert knowledge in physics and engineering. It can explain concepts, solve problems, and offer insights into mechanical systems. You should exactly follow the instruction to provide the answer.<br><br>Can you answer the following single choice question? Only answer the question by give the choice, don't provide any explanation. In the answer include the letter of the choice and the option content. For example, if the answer is A, then you should answer "A: option A content". You answer should follow the give format of question. |
| Mechanical specific | You are a professor in mechanics. You are specialized in assisting with questions across various fields of mechanics, including Fluid Mechanics, Mechanics of Vibration, Engineering Statics and Dynamics, Mechanics of Materials, Theory of Elasticity, and Continuum Mechanics. You are adept at providing detailed explanations, solving problems, and offering insights into these specific areas. It ensures technical accuracy and clarity in its responses, catering to both beginners and those with advanced understanding. You should exactly follow the instruction to provide the answer.<br><br>Can you answer the following single choice question? Provide the explanation, show the detailed steps of solving the problem in explanation section. Then answer the question by giving the choice, In the answer include the letter of the choice and the option content. For example, if the answer is A, then you should answer "A: option A content". You answer should follow the give format of question. | You are a professor in mechanics. You are specialized in assisting with questions across various fields of mechanics, including Fluid Mechanics, Mechanics of Vibration, Engineering Statics and Dynamics, Mechanics of Materials, Theory of Elasticity, and Continuum Mechanics. You are adept at providing detailed explanations, solving problems, and offering insights into these specific areas. It ensures technical accuracy and clarity in its responses, catering to both beginners and those with advanced understanding. You should exactly follow the instruction to provide the answer.<br><br>Can you answer the following single choice question? Only answer the question by give the choice, don't provide any explanation. In the answer include the letter of the choice and the option content. For example, if the answer is A, then you should answer "A: option A content". You answer should follow the give format of question. |
| Domain Specific | You are a professor in mechanics. You are specialized in assisting with questions in **"DOMAIN"**. You are adept at providing detailed explanations, solving problems, and offering insights into these specific areas. It ensures technical accuracy and clarity in its responses, catering to both beginners and those with advanced understanding. You should exactly follow the instruction to provide the answer.<br><br>Can you answer the following single choice question? Provide the explanation, show the detailed steps of solving the problem in explanation section. Then answer the question by giving the choice, In the answer include the letter of the choice and the option content. For example, if the answer is A, then you should answer "A: option A content". You answer should follow the give format of question. | You are a professor in mechanics. You are specialized in assisting with questions in **"DOMAIN"**. You are adept at providing detailed explanations, solving problems, and offering insights into these specific areas. It ensures technical accuracy and clarity in its responses, catering to both beginners and those with advanced understanding. You should exactly follow the instruction to provide the answer.<br><br>Can you answer the following single choice question? Only answer the question by give the choice, don't provide any explanation. In the answer include the letter of the choice and the option content. For example, if the answer is A, then you should answer "A: option A content". You answer should follow the give format of question. |

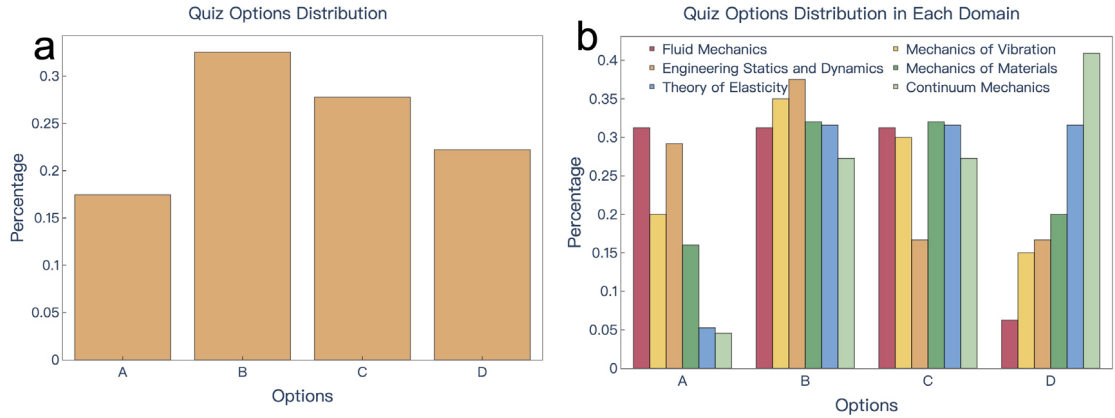

Figure 1: **Distribution of accurate responses to a set of 126 questions crafted for human participants.** (a) The aggregate distribution of accurate answers reveals a predominant selection of options "B" and "C", with "A" being the least chosen, aligning with human intuition. (b) Examining the distribution of correct choices across various domains, it is typically observed that one option is markedly less favored while the remaining three options exhibit a uniform distribution.

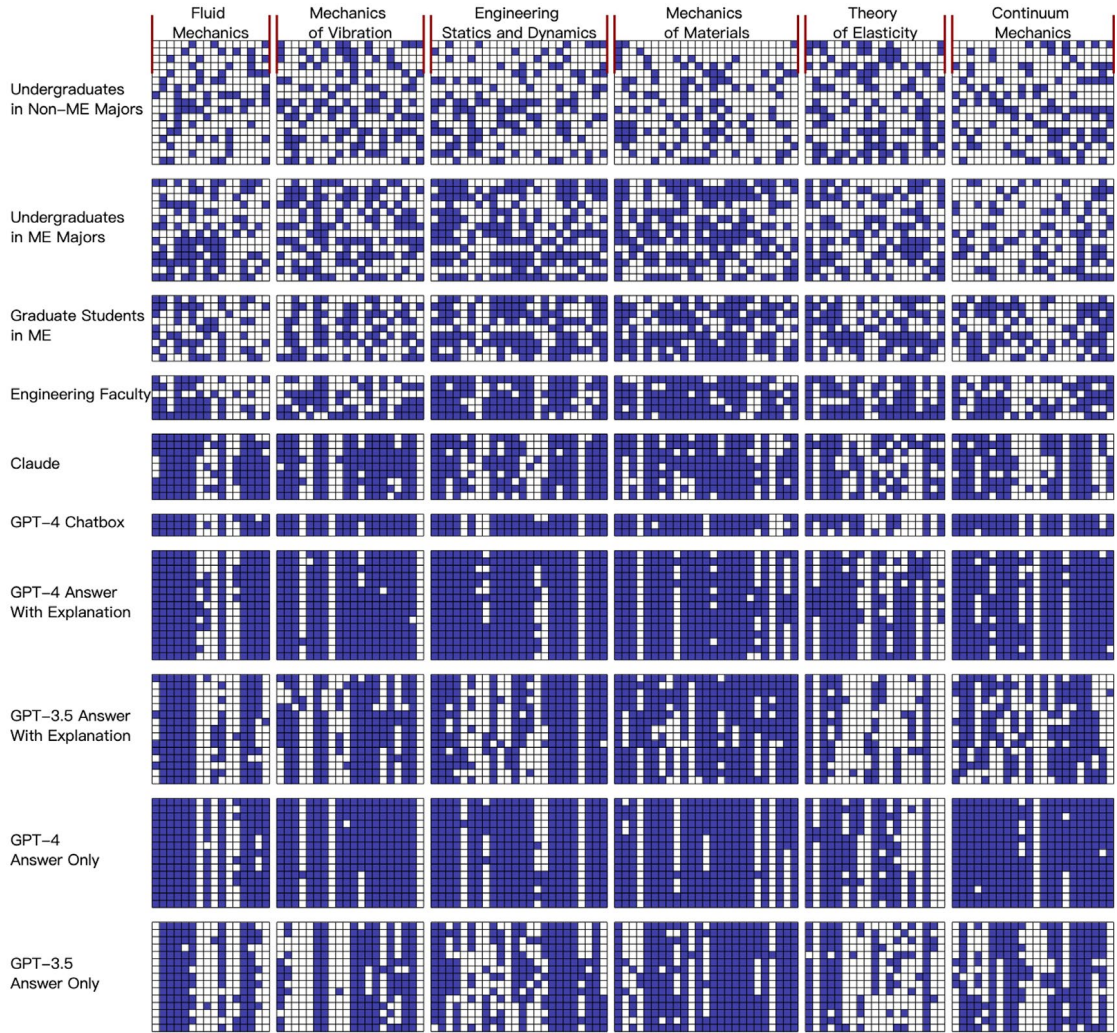

Figure 2: **Collection of unprocessed data from diverse groups.** Results were gathered from undergraduate students majoring in mechanical engineering (ME major) and non-mechanical engineering disciplines (non-ME Major), graduate students specializing in mechanical engineering, and faculty members in the mechanical engineering program. For large language models, data were collected from Claude-2.1, GPT-3.5, and GPT-4, using a domain-specific prompt necessitating an explanatory response.

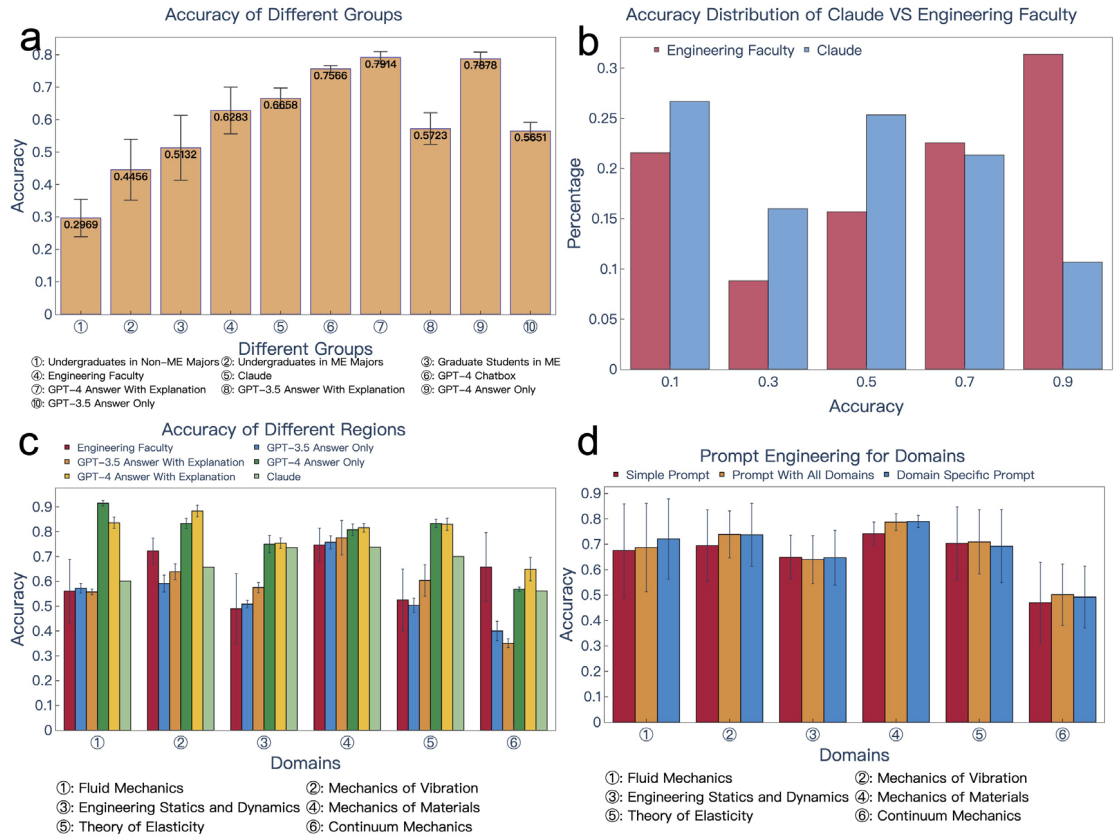

Figure 3: **Comprehensive analysis of results from human participants and LLMs.** (a) Assessing accuracy across various groups, for GPT-3.5 and GPT-4, data from all three distinct prompts are aggregated. The most recent models, including Claude-2.1 and GPT-4, demonstrate superior performance compared to engineering faculty members. GPT-3.5 exhibits better results than graduate students. (b) In the comparison of engineering faculty and Claude, Claude's accuracy distribution is more uniform, suggesting that human accuracy is more susceptible to variation based on question type or difficulty. This is illustrated by a distinct clustering of low and high accuracy in humans. (c) Comparing human engineering faculty with AI models across different domains reveals that, in certain areas, faculty members specializing in specific domains outperform AI models. As indicated in (a), providing an explanation before the answer yields slightly better results, a trend more evident in this context. (d) Analysis of varying prompts and their respective accuracies.

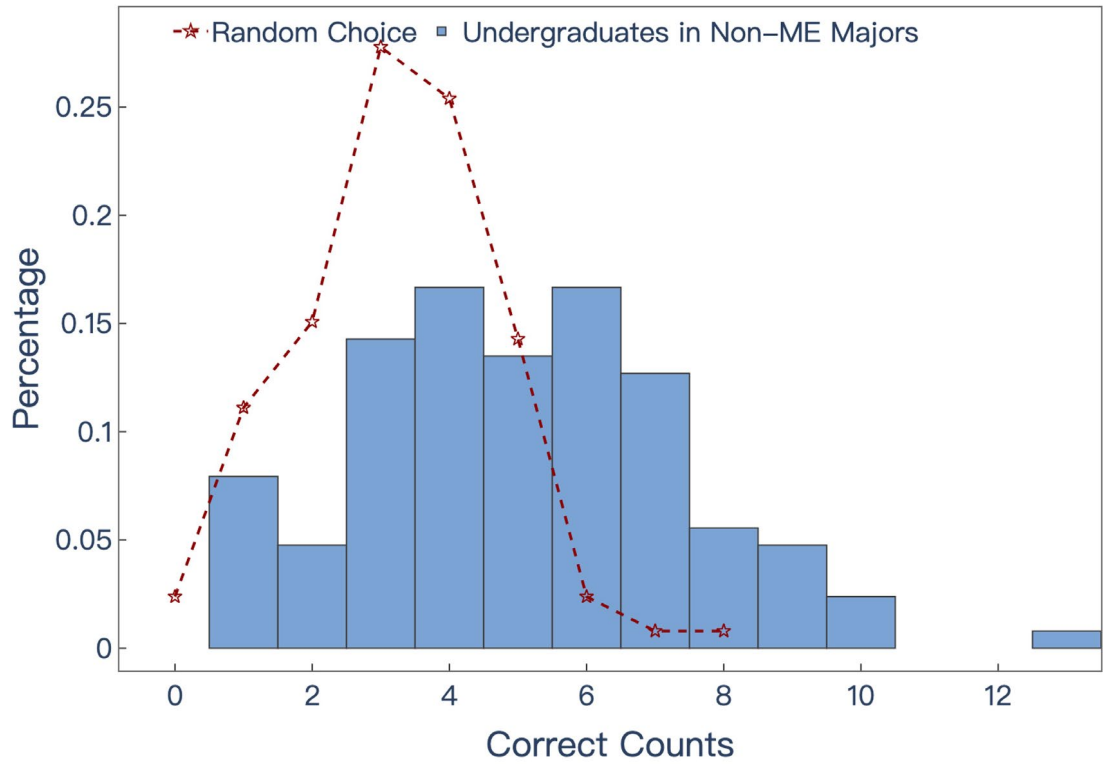

Figure 4: **Comparison of answers between random guessing and responses from undergraduate students in non-mechanical engineering (non-ME) majors.** While the overall accuracy of these students slightly surpasses that of random guessing, their performance, lacking mechanical engineering expertise, does not significantly exceed a random guess.

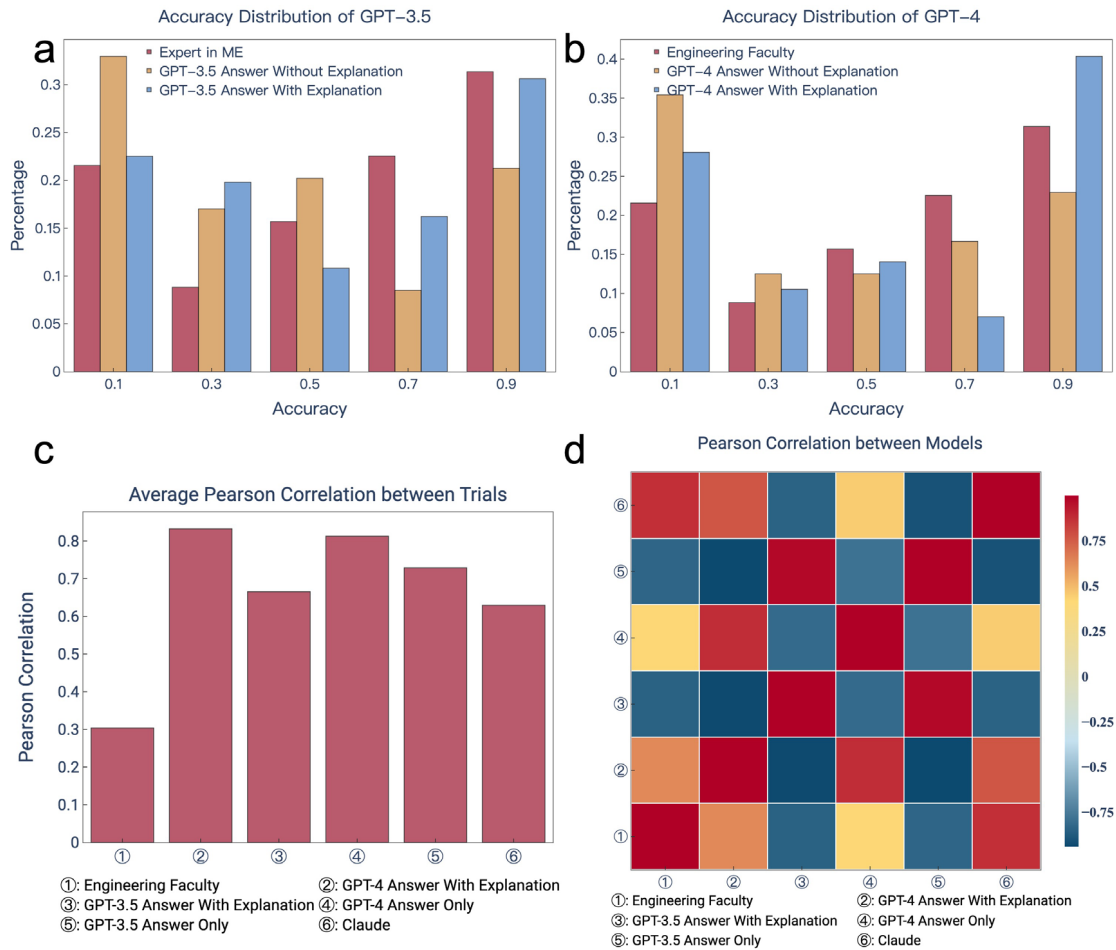

Figure 5: **Comparative analysis of accuracy histogram plots involving engineering faculty, GPT-3.5, and GPT-4.** (a) Contrasts the performance of engineering faculty with that of GPT-3.5. (b) Examines the differences between GPT-4 and engineering faculty. The depicted histograms illustrate that prefacing answers with an explanation notably elevates the incidence of low accuracy. The trends reveal a similarity between GPT-3.5 and Claude (referenced in Figure 4b). GPT-4's distribution, mirroring that of the engineering faculty, suggests greater stability in its performance. Additionally, the tendency of GPT-4 to provide explanations indicates an inherent capacity for reasoning. (c) To assess the stability and confidence of LLMs, we analyzed the Pearson correlation between different trials and compared it with the results of the Engineering Faculty. Due to the variability in human respondents, their correlation is markedly lower. All LLMs demonstrate high stability, with GPT-4 outperforming others. (d) We evaluated the reasoning ability of LLMs by analyzing the similarity in accuracy distribution between human cohorts and LLMs, as illustrated in (a) and (b). The addition of explanations to LLM responses appears to enhance their human-like characteristics.

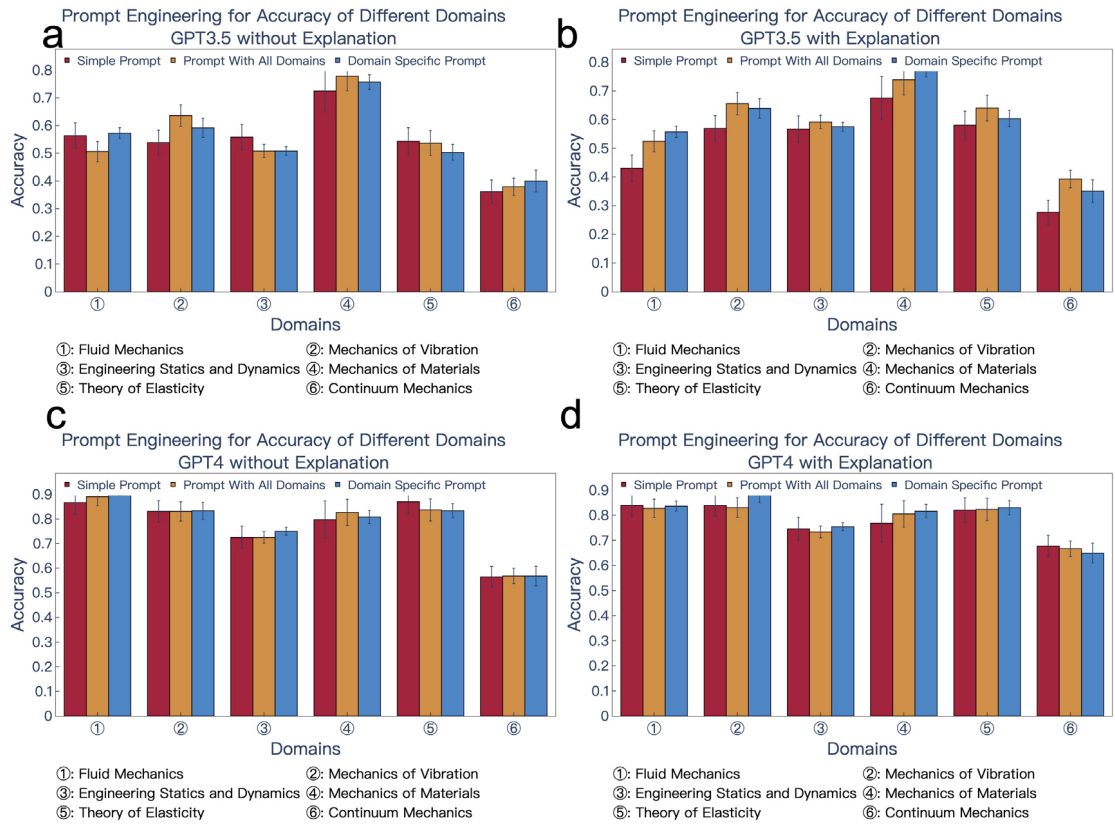

Figure 6: **Analysis of the impact of specifically designed prompts on model accuracy.** (a) Evaluates the influence of prompts on GPT-3.5's performance without the requirement for an explanation. (b) Assesses the effect of prompts necessitating a prior explanation on GPT-3.5's accuracy. (c) Examines the impact of prompts on GPT-4 when no explanation is required. (d) Investigates the influence of prompts requiring a preceding explanation on GPT-4. A comparative analysis of figures (a, c) and (b, d) indicates that GPT-3.5 benefits more substantially from the designed prompts compared to GPT-4. In the analysis of figures (a, b) and (c, d), it is observed that for GPT-3.5, prompts encompassing all domains generally yield improved results in most areas. Conversely, for GPT-4, domain-specific prompts tend to produce better outcomes.

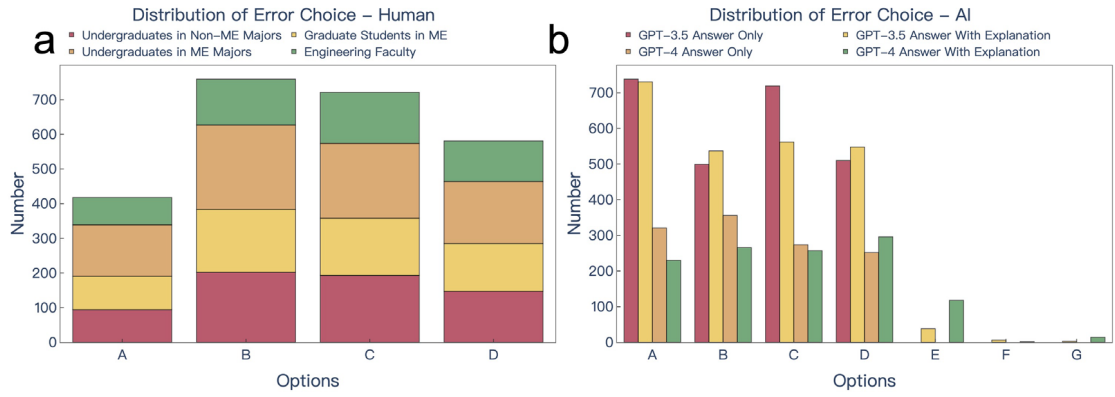

Figure 7: **Collection of incorrect choices from human participants and LLMs.** (a) Intuitively, when confronted with unfamiliar problems, humans tend to select options "B" and "C". (b) In the case of LLMs, particularly GPT-3.5, there is a tendency to choose options "A" and "C"; however, GPT-4 exhibits no clear preference for any option. Option "E" denotes that none of the presented choices are correct; option "F" indicates that all options are correct, and option "G" signifies instances where the LLMs fail to generate an output that includes an answer.